\documentclass[10pt, a4paper]{article}
\usepackage{lrec2000}
\usepackage{alltt,epsfig}

\def\smtt#1{{\small\tt #1}}

\title{Many Uses, Many Annotations for Large Speech Corpora: \\
Switchboard and TDT as Case Studies}

\name{David Graff and Steven Bird}

\address{Linguistic Data Consortium \\
	University of Pennsylvania \\
	3615 Market Street, Philadelphia, PA 19104, USA \\
	{\small\tt http://www.ldc.upenn.edu}}

\abstract{This paper discusses the challenges that arise when large
speech corpora receive an ever-broadening range of diverse and
distinct annotations. Two case studies of this process are
presented: the Switchboard Corpus of telephone conversations
and the TDT2 corpus of broadcast news.  Switchboard
has undergone two independent transcriptions and various types of
additional annotation, all carried out as separate projects that were
dispersed both geographically and chronologically.  The TDT2 corpus
has also received a variety of annotations, but all
directly created or managed by a core group.  In both cases, issues
arise involving the propagation of repairs, consistency of references,
and the ability to integrate annotations having different formats and
levels of detail.  We describe a general framework whereby these
issues can be addressed successfully.}

\begin{document}

\maketitleabstract

\section{Introduction}

Any well-constructed corpus of speech data can provide a valuable
resource for a wide variety of uses in language research and
technology development, especially if the corpus is centered on common
and naturally-occurring speech events.  Both the potential and the
motivation for re-use increase with the size of the corpus: larger
corpora provide a better representation of linguistic diversity and
variability, and so are richer objects of study for any research goal;
also, the expense and effort that go into the creation of a large corpus,
typically on behalf of a particular research program, can provide
powerful leverage for researchers involved in other projects or areas
of study.

It is often the case that a new use of a corpus will require a new
annotation that was not part of the initial corpus creation effort.
But for large corpora, it often happens that the new annotations can
be applied only to subsets of the collection, depending on the
difficulty of the new task and the resources available to carry it
out.  In this case, it becomes increasingly important, and
increasingly difficult, to maintain the coherence of shared resources.

The Linguistic Data Consortium (LDC) has been involved in managing
multiple annotations of two large speech corpora: the Switchboard
Corpus of conversational telephone speech (SWB), and the Topic
Detection and Tracking corpus of broadcast news (TDT).  In the case of
SWB, the original corpus creation and all subsequent annotations have
been done by others outside the LDC, and we have acted simply as the
central point of contact and access for all users and annotators; in
the case of TDT, LDC personnel have been responsible for most
forms of annotation done so far.

The following sections will review these corpora in
terms of their overall content, the particular types of annotations
that have been applied to them so far, and the relative coverage of
those annotations (including the overlap that exists among them).  We
then discuss some of the problems that arose due to localized data
repairs that were applied in some annotations, and measure the extent
of referential consistency among the diverse annotations.  Finally, we
describe how the application of annotation graph structure to the
various derivative data sets provides a powerful and flexible means
for comparing and integrating their contents.

\section{The Switchboard Corpus}

Switchboard-1, the first large collection of spontaneous
conversational speech over the telephone, was collected in 1990 by
Texas Instruments (TI) \cite{Godfrey}. The current published corpus
comprises 2438 calls involving 520 native speakers of American
English, recruited from all over the United States.  The calls range
between five and ten minutes in length, and each call contains a
unique pairing of speakers.  Individual speakers are designated by
identification numbers, and information is provided as to their
gender, age, education and dialect region.  On average, each speaker
appears in about nine calls (the range of calls per speaker varies
between 1 and 32).

The speech data for each call is provided in the form of a two channel
interleaved sample file; the two channels, designated ``A'' and ``B'',
represent the mu-law encoded signal received from the telephone
handsets of the two speakers participating in the call.

\subsection{Initial transcription}

TI created an initial set of manual transcripts, employing
professionally trained transcribers equipped with analog tape copies
of the sample files.  These transcripts were then submitted to an
automatic speech recognition (ASR) process to establish approximate
time alignments at the word level.

The initial publication of the corpus in 1993 contained 2288 calls on
25 CD-ROMS (additional calls were being held in reserve at that time
for future use as test data).  A separate CD-ROM provided the
transcripts in two forms (two files per call): (1) the original text
files as created by transcribers, including a generic header with
information about the call, and speaker designations (``A'' or ``B'')
for each turn in the dialog, but no time annotations; (2) the
time-aligned version of the text, with one word token per line,
accompanied by the speaker designation, and the beginning time and
duration of the word.  The transcript release also included
documentation and tables describing the corpus content.

Given the size, complexity and novelty of this effort, a variety of
problems affected the first release of the corpus by the LDC:

\begin{itemize}

\item
In about 200 calls, the transcribers had made mistakes in assigning
the speaker labels (``A'' or ``B'') to some or all of the turns in the
transcripts.  Because the speaker label is intended to represent the
signal channel in the speech file, an error in the speaker label would
direct the corpus user to the wrong channel when retrieving the audio
data for a given turn in the dialog.

\item
Other more typical kinds of errors, involving misperceptions by
transcribers about what was said, presumably affected nearly all
files, but only to a very small degree.

\item
The ASR word alignment was applied to a single-channel (combined)
version of the speech data, and so was not affected by speaker label
errors, but it was relatively ineffective in regions where turns
overlapped in time, and where the transcript contained omissions,
additions or substitutions of words relative to what had been spoken.

\item
In about 30 calls, the two forms of transcript data did not
match up in terms of the number of speaker turns present in the
dialog.

\item
For about 30 additional calls, some or all of the transcript data were
absent from the publication, because the text files were missing,
incomplete or corrupted due to errors in preparing the publication.

\end{itemize}

In 1994, the LDC released an ``interim'' version of the transcripts,
fixing all cases of the last two types of problems, and about one quarter of
the files containing known speaker-label errors (particularly, those
files where the label errors were limited to small portions of the
transcript).  The remaining speaker-label problems were documented as
well as possible, but were not fixed, nor was any attempt made to
correct other transcription or time-alignment errors.

\subsection{Subsequent annotations} 

Since the initial release of SWB transcripts, a number of research
sites have used this data, either with or without reference to the
speech, as input to a range of divergent annotation projects.  These
are briefly described below, in roughly chronological order.  Some
of the resulting annotations are displayed in Figure~\ref{fig:swb}.

\begin{figure*}
{\tiny\setlength{\tabcolsep}{.25\tabcolsep}
\begin{tabular}{l|l|l}
\begin{minipage}[t]{.21\linewidth}
{\small\bf Aligned Word}
\begin{alltt}
B 19.44 0.16 Yeah,
B 19.60 0.10 no
B 19.70 0.10 one
B 19.80 0.24 seems
B 20.04 0.02 to
B 20.06 0.12 be
B 20.18 0.50 adopting
B 20.68 0.16 it.
B 21.86 0.26 Metric
B 22.12 0.26 system,
B 22.38 0.18 no
B 22.56 0.06 one's
B 22.86 0.32 very,
B 23.88 0.14 uh,
B 24.02 0.16 no
B 24.18 0.32 one
B 24.52 0.28 wants
B 24.80 0.06 it
B 24.86 0.12 at
B 24.98 0.22 all
B 25.66 0.22 seems
B 25.88 0.22 like.
A 28.44 0.28 Uh,
A 29.26 0.14 the,
A 29.48 0.14 the,
A 29.82 0.10 the
A 29.92 0.34 public
A 30.26 0.06 is
A 30.32 0.22 just
A 30.54 0.14 very
A 30.68 0.68 conservative
A 31.36 0.18 that
A 31.54 0.30 way
A 32.56 0.12 in
A 32.74 0.64 refusing
A 33.60 0.12 to
A 33.72 0.56 change
A 34.94 0.48 measurement
A 35.42 0.62 systems,
A 36.08 0.26 uh,
A 37.04 0.38 money,
A 37.62 0.30 dollar,
A 37.92 0.46 coins,
A 38.38 0.22 anything
A 38.60 0.18 like
A 38.78 0.30 that.
B 39.34 0.10 Yeah
B     *    * [laughter].
A 40.96 0.04 And,
A 41.32 0.04 and,
A 42.28 0.36 and
A 42.88 0.20 it
A     *    * [breathing],
A 43.08 0.16 it
A 43.48 0.46 obviously
A 43.94 0.22 makes
A 44.16 0.14 no
A 44.30 0.36 sense
A 44.66 0.06 that
A 44.72 0.12 we're
A 44.84 0.70 practically
A 46.52 0.32 alone
A 46.84 0.10 in
A 46.94 0.06 the
A 47.00 0.44 world
A 47.44 0.16 in,
A 48.52 0.04 in
A 48.56 0.26 using
A 48.82 0.08 the
A 48.90 0.22 old
A 49.12 0.40 system.
\end{alltt}
\end{minipage}
&
\begin{minipage}[t]{.2\linewidth}
{\small\bf Part of Speech}
\begin{alltt}
====================
[ SpeakerB22/SYM ]
./. 
====================

Yeah/UH ,/, 
[ no/DT one/NN ]
seems/VBZ to/TO
be/VB adopting/VBG 
[ it/PRP ] ./. 

[ Metric/JJ system/NN ]
,/, 
[ no/DT one/NN ]
's/BES very/RB ,/, 
[ uh/UH ] ,/, 
[ no/DT one/NN ]
wants/VBZ 
[ it/PRP ]
at/IN 
[ all/DT ]
seems/VBZ like/IN ./. 

====================
[ SpeakerA23/SYM ]
./. 
====================

[ Uh/UH ] ,/, 
[ the/DT ] ,/, 
[ the/DT ] ,/, 
[ the/DT public/NN ]
is/VBZ just/RB very/RB
conservative/JJ that/DT 
[ way/NN ]
in/IN refusing/VBG
to/TO change/VB 
[ measurement/NN
  systems/NNS ]
,/, 
[ uh/UH ] ,/, 
[ money/NN ] ,/, 
[ dollar/NN ] ,/, 
[ coins/NNS ] ,/, 
[ anything/NN ]
like/IN 
[ that/DT ] ./. 

====================
[ SpeakerB24/SYM ]
./. 
====================

Yeah/UH ./. 

====================
[ SpeakerA25/SYM ]
./. 
====================

And/CC ,/, and/CC ,/,
and/CC 
[ it/PRP ] ,/, 
[ it/PRP ]
obviously/RB makes/VBZ 
[ no/DT sense/NN ]
that/IN 
[ we/PRP ]
're/VBP practically/RB
alone/RB in/IN 
[ the/DT world/NN ]
in/IN ,/, in/IN
using/VBG 
[ the/DT old/JJ
  system/NN ]
./. 
\end{alltt}
\end{minipage}
&
\begin{minipage}[t]{.5\linewidth}
{\small\bf Disfluency}
\begin{alltt}
B.22:   Yeah, / no one seems to be adopting it. /
  Metric system, [ no one's very, + {F uh, } no one wants ]
  it at all seems like. / 
A.23:   {F Uh, } [ [ the, + the, ] + the ]
  public is just very conservative that way in
  refusing to change measurement systems,
  {F uh, } money, dollar, coins, anything like that. /
B.24:   Yeah <laughter>. /
A.25:   [ [ {C And, } +  {C and, } ] + {C and } ]
  [ it + <breathing>,  it ] obviously makes no sense
  that we're practically alone in the world [ in, + in ]
  using the old system. /
\end{alltt}

{\small\bf Treebank}
\begin{alltt}
((CODE SpeakerB22 .))
((INTJ Yeah , E_S))
((S (NP-SBJ-1 no one)
    (VP seems
        (S (NP-SBJ *-1) 
           (VP to (VP be (VP adopting (NP it)))))) . E_S))
((S (NP-TPC Metric system) ,
    (S-TPC-1 (EDITED (RM [)
                     (S (NP-SBJ no one)
                        (VP 's (ADJP-PRD-UNF very))) ,
                     (IP +)) (INTJ uh) ,
             (NP-SBJ no one)
             (VP wants (RS ]) (NP it) (ADVP at all)))
    (NP-SBJ *)
    (VP seems (SBAR like (S *T*-1))) . E_S))
((CODE SpeakerA23 .))
((S (INTJ Uh) ,
    (EDITED (RM [)
            (EDITED (RM [) (NP-SBJ-UNF the) , (IP +))
            (NP-SBJ-UNF the) , (RS ]) (IP +))
    (NP-SBJ-1 the (RS ]) public)
    (VP is
        (ADVP just)
        (ADJP-PRD very conservative)
        (NP-MNR that way)
        (PP in
            (S-NOM (NP-SBJ-2 *-1)
                   (VP refusing
                       (S (NP-SBJ *-2)
                          (VP to
                              (VP change
                                  (NP (NP measurement systems) ,
                                      (INTJ uh) , (NP money) ,
                                      (NP dollar) , (NP coins) ,
                                      (NP (NP anything)
                                          (PP like
                                              (NP that))))))))))) . E_S))
((CODE SpeakerB24 .))
((INTJ Yeah . E_S))
((CODE SpeakerA25 .))
((S (EDITED (RM [)
            (EDITED (RM [) And , (IP +)) and , (RS ]) (IP +)) and (RS ])
    (EDITED (RM [) (NP-SBJ it) (IP +) ,)
    (NP-SBJ (NP it)
            (SBAR *EXP*-1))
    (RS ])
    (ADVP obviously)
    (VP makes
        (NP no sense)
        (SBAR-1 that
                (S (NP-SBJ-2 we)
                   (VP 're
                       (ADVP practically) (ADJP-PRD alone)
                       (PP-LOC in (NP the world))
                       (EDITED (RM [) (PP-UNF in) , (IP +))
                       (PP in (RS ])
                           (S-NOM (NP-SBJ *-2)
                                  (VP using
                                      (NP the old system)))))))) . E_S))
\end{alltt}
\end{minipage}
\end{tabular}}
\vspace*{2ex}\hrule

\caption{Multiple Annotations of the Switchboard Corpus}\label{fig:swb}
\end{figure*}

\subsubsection*{Phrase-level time stamps (BBN)}

The first application of SWB data for research in large vocabulary
conversational speech recognition was in conjunction with the DARPA
LVCSR project.  In preparation for initial training on SWB,
researchers at BBN created a modified version of the original TI
transcripts, and circulated this among the LVCSR participants.  The
modification involved forming time-stamped ``phrasal'' regions from the word-level
time-aligned transcripts, and assigning a unique identifier to
each region.  This produced a segmentation of the dialogs that differed
from the turn units created by the original transcribers: a single
time-stamped phrase might encompass parts of two consecutive turns, if
the original transcriber had broken up the phrase in order to insert
an interruption by the other speaker; also, multiple time-stamped
phrases might be derived from a single turn, if that turn was
considered too long for reliable ASR processing.  The speaker-label
errors mentioned above were not repaired in this process (though
a list of affected files was circulated among LVCSR participants).
This form of the transcripts has never been made available through the
LDC.

\subsubsection*{Disfluency annotation (Penn Treebank Project)}

As a preliminary step to prepare for Treebank annotation, 650
transcript files were selected from the LDC ``interim'' release and
annotated for various types of
disfluency that occurred in the spontaneous speech.  This annotation
was essentially text-based, relying entirely on the representation of
disfluencies in the (non-time-aligned) transcripts
(see Figure~\ref{fig:swb}).  The objective was
to tag hesitations, stutters, word fragments, restarts, and a limited
class of discourse markers, since these elements in the transcripts
would constitute exceptions or barriers to syntactic analysis.  These
annotations are included in the current ``Treebank 3'' corpus,
available from the LDC.

\subsubsection*{POS-tagging, parsing (Penn Treebank Project)}

Building on the output of the disfluency annotations, the Penn
Treebank Project applied part-of-speech tagging and syntactic parsing
to the 650 files that had been selected (see Figure~\ref{fig:swb}).
In the course of this
annotation, a small number of corrections were made to the text data,
consisting mostly of repairs to punctuation and replacements of some
incorrect words.  Again, this annotation was done without reference to
the audio data.  It is currently available as the ``Treebank 3''
corpus.

\subsubsection*{Discourse annotation (Univ. of Colorado, SRI)}

This annotation project set out to ``model the speech act type of each
utterance'' and ``model sociolinguistic facts about conversation
structure...'' \cite{Jurafsky} It builds on the disfluency annotation
mentioned above, using the same conceptual basis for segmenting dialog
turns into phrases (``utterances'') that are cohesive in terms of the
speech acts being performed.  An exhaustive segmentation of such
utterances was carried out on 1155 conversations, and each utterance
was categorized as to speech act or discourse function.  All 650 files
covered by the Treebank Project are included in this set.  The data is
available directly from the University of Colorado \cite{Jurafweb}.

\subsubsection*{Phonetic transcription (ICSI)}

The International Computer Science Institute (ICSI) at the University
of California, Berkeley, began a project in 1996 to carry out
fine-grained phonetic annotations on portions of SWB data \cite{icsi}.
The selection of portions to transcribe, as well as the initial
orthographic transcriptions, were apparently derived from the BBN
phrase-level segmentation, and the resulting annotations are indexed
by means of the unique identifier strings assigned to each phrase by
BBN.  In contrast to the other annotation projects, ICSI selected a
sampling of 5100 phrases from a wide range of files, with each phrase
ranging from 0.45 to 17.430 seconds (the majority are between 3 and 5
seconds).  Two files received fairly exhaustive treatment for one side
of the call, but for 1602 other files, the typical coverage of ICSI
transcriptions is on the order of tens of seconds.  Altogether, the
combined annotations cover nearly 3.5 hours of speech.  This data set
is currently available from the Center for Language and Speech
Processing at Johns Hopkins University \cite{clsp}.

The initial stage of the project assigned time marks at the level of
segmental boundaries, as well as syllable and word boundaries.  In the
second phase, time boundaries were applied at the word and
syllable levels.  Throughout the project, the labels assigned to
phonetic segments (whether time marked individually or at the syllable
level) were intended to accurately reflect the actual pronunciation in
the signal, to a much finer level of detail than in previous corpora
(e.g. TIMIT or the Boston University Radio corpus).

\subsubsection*{Complete resegmentation (ISIP)}

In view of the importance of SWB as a multi-functional corpus, and
the difficulties that have accompanied the original transcripts, an
important effort was launched at the Institute for Signal and
Information Processing (ISIP) at Mississippi State University, to
conduct a complete review of the transcription data, applying a new
speech transcription tool developed at ISIP with particular attention
to locating known types of problems and avoiding known pitfalls in
this type of annotation effort.  By the time this project began, the
SWB calls that had previously been held back as test data had been
used over the course of several benchmark tests in the LVCSR project,
and were now available for publication.  The current release of SWB
speech and transcript data now comprises 2438 calls.  By agreement
with the LDC, ISIP has made the complete set of SWB transcripts freely
available on their web site \cite{isipswb}.

\subsection{Rectification and Integration}

With the completion of the ISIP review of SWB transcripts, it is now
possible to assess the impact of transcription errors on the various
divergent annotations.  As an initial step to check for the magnitude
of errors involving lexical content, we treated the LDC ``interim''
transcripts as a test set to be measured for error rates, using the
ISIP transcripts as the reference text.  Comparable versions of the
two data sets were constructed by aligning the phrase-level ISIP time
marks with the original TI word-level time marks, and the NIST scoring
tool ``SCLite'' was used to calculate insertions, deletions and
substitutions. The results are summarized in table \ref{wertable}.
\footnote{LDC files that still contained speaker/channel labeling
errors over some or all of their transcripts were not included in
this scoring.}  It should be noted that word fragments and non-lexical
tokens (e.g. ``uh-hum'' vs. ``uh-huh'') accounted for roughly 30\%
(over 19,000) of the substitution errors, and about 21\% (nearly
34,000) of the insertion and deletion errors.

\begin{table}[ht]
\begin{center}
\begin{tabular}{|r|c|r|r|c|}

      \hline
      Units     & Status & K     & \% & Per-file \% range \\
      \hline\hline
      phrases   & correct & 136 & 55.1 &  \\
		& w/errors & 111 & 44.9 &  7.6 - 90.9 \\
      \hline
      words     & correct & 2895 & 94.8 & 77.4 -  99.5 \\
		& accuracy &    & 92.8 & \\
                & all errors & 220 & 7.2 & 0.8 - 27.9 \\
      		& substit. & 63 & 2.1 & 0.0 - 14.5 \\
                & deleted & 95  & 3.1 & 0.3 - 15.5 \\
                & inserted & 63 & 2.0 & 0.0 - 21.0 \\
      \hline

\end{tabular}
\end{center}
\caption{Summary of word errors in LDC ``interim'' transcripts} 
\label{wertable}
\end{table}

The comparison of ICSI annotations to the other versions of SWB
transcripts is somewhat more problematic.  ICSI transcribers made
corrections to the lexical content in accordance with their more
detailed attention to the actual pronunciation of phrases, but they
used different conventions regarding word hyphenation and
disfluencies, and occasionally inserted annotations for non-linguistic
events (e.g. ``breath'', ``mouthnoise'', etc.) without the intended
markup to distinguish these from lexical tokens.  Still, we can
estimate the upper bound on the number of corrections imposed by ICSI,
again using the NIST SCLite scoring method.  In this case, the LDC
transcripts contained 96\% of the words in the 3.5 hours of ICSI
word-level data -- there were, at most, 2\% omissions and 2\%
substitutions in the LDC transcripts.

ICSI's use of the BBN phrasal time marks as the basis of phrase
selection creates some additional difficulties: 

\begin{itemize}

\item
Correlation with original TI word-level time marks is imperfect at
best; insertion errors predominated in scoring the LDC transcripts
(6.2\%, three times more than deletions or substitutions), yielding an
overall word accuracy of 89.2\%.  This was due mostly to discrepancies
at phrase boundaries.

\item
Some of the BBN ``phrase boundaries'' occur at impractical positions
within syllables, causing some boundary tokens to be interpreted
differently.

\item
Until a reliable word-level time marking is done for the ISIP
transcripts, there will be no reasonable way to align the ICSI and
ISIP annotations, due to significant differences in phrase
segmentation.

\end{itemize}

Overall, these tabulations indicate that the lexical accuracy of the
original TI transcripts was quite high, and the impact of word errors
on downstream annotations, particularly the Treebank and discourse
data, may be considered negligible.

A major challenge remains, however, in terms of integrating the
various annotations.  All three versions of time-marked transcripts
(LDC ``interim'', ICSI and ISIP) assign unique identifiers to each
turn or phrase unit, but each set has a distinct inventory of units
that cannot, as yet, be cross-referenced to the other two sets in any
reliable way.  The only stable point of reference is the audio data,
and the use of time offsets into the speech files.

\section{The TDT Corpora}

The design and content of the TDT corpora are described in other
presentations at LREC-2000 \cite{wayne,cieri}.  The present discussion
will focus on the range of distinct annotations applied to the data,
the relationships among them, and the problems involved in
coordinating them.

\subsection{Multiple data streams from audio and text sources}

The audio recordings of broadcast (video and radio) sources in TDT
were used to create a variety of textual data streams.  For video
sources (which have all been in English), the broadcast signal
included closed-caption text data, which was converted to
computer-readable form while the audio was being digitized; also, one
video source (ABC News) provided full transcripts of its daily
broadcasts to the public through a commercial transcription service.
It is well known that closed-caption text tends to be incomplete
relative to what is actually spoken during a broadcast, because the
maximum practical display rate, in words per minute, is slower than
typical rates of speech.  The full transcripts produced by commercial
services, which are intended for use as a standard public record of
broadcast content, are lexically correct to a high degree of accuracy,
though they typically avoid the inclusion of any disfluencies -- the
text represents only what the speaker {\em intended} to say, and omits
filled pauses, false starts, stutters, and the like.  On the basis of
this one source, then, it is possible to estimate a baseline of ``word
error rate'' for closed-caption text, which can be useful when
assessing other sources for which only the closed-captions are
available.

The radio sources in TDT did not have publicly available transcripts,
and four different transcription services were enlisted to transcribe
these programs as part of the corpus creation effort (one for all TDT
Mandarin data, one for TDT3 English, and two others for TDT2 English).
The services varied in terms of the quality of transcripts delivered,
with one of the TDT2 English services having been poorest.

In addition to manual transcription, all audio sources were submitted
to unguided ASR, to establish a benchmark of TDT system performance
given this quality of text as input.  The TDT2 English data was
submitted to two different English-based ASR systems; of course, in
the absence of fully accurate manual transcriptions for most of this
material, it remains difficult to compare their performance in terms
of word error rate.

For all Mandarin sources, both newswire text and radio transcripts,
additional data streams were produced by running the data through a
Chinese-to-English machine translation system (SYSTRAN), with no
manual guidance or revision of the system output.  For audio data,
both the manual and ASR transcriptions were translated in this way.
Again, this was intended to set a benchmark for cross-lingual TDT
performance given this quality of input.

As a result, most data sources were represented by at least two
parallel, independent data streams -- a couple of sources had three or
four streams -- each with its own peculiar properties and token
sequence.  The stable points of reference across all streams were the
boundaries between news stories, and in the case of audio sources, the
time offsets of those boundaries in the speech data.

\subsection{Creating and tracking multiple annotations}

A relational database was used to track the main stages of data
creation and TDT-specific annotations.  The basic units of the corpus
are the sample file and the topical story unit.  For audio sources, an
entry was created each time a recording process was scheduled; the
entry was updated at the conclusion of the recording, updated again
after manual inspection to determine whether the recording was
successful, and again after manual segmentation of the 30- or
60-minute file into story and non-news segments.  This done, each
story was assigned a unique identifier, which included the source,
date, broadcast start time, and time offset within the file at the
start of the story, and these identifiers were entered into the
database to guide the topic annotation.  (The equivalent stages for
newswire data were fully automated to prepare the stream or bulk
archive text data for topic annotation.)

During the main topic annotation phase, in which every story had to be
read some minimum number of times to assess it against all selected
TDT topics, annotators also had to decide whether a given story was
flawed in any of four ways, making it unsuitable for topic labeling.
In the newswire data, a reported flaw would typically result in the
removal of a story from the corpus, but for broadcast data, a flaw
would generally be the result of a mistake during the manual story
segmentation phase.  Broadcast stories could not simply be discarded,
as this would create gaps in the coverage of the continuous audio
signal -- segmentation errors needed to be fixed, and this would
affect the inventory of news stories in the file, and/or the locations
of boundaries (hence the identification numbers assigned to the
stories would change, as well).

A further complication was the need in TDT3 to support alternative
methods of topic annotation while the main annotation was still in
progress.  These alternative forms of labeling -- first-story detection
and story-link detection -- were not actively tied into the database
management of the main topic annotation; rather, they used a snap-shot
of the corpus, taken as late and as carefully as possible during the
main annotation.  Fortunately, they would involve only a subset of the
full TDT3 corpus, so it was possible to avoid particular files or
stories where problems had been observed, and still provide an adequate
sample.  Despite our best efforts, some of the results of these
alternate annotations could not be used in the final delivery of the
corpus, because the stories to which they applied had been altered or
removed in the course of repairs prompted by the main annotation; in
particular, 0.5\% of the 21,600 story link annotations were discarded
because a number of stories had been eliminated from the corpus.

Additional uses of TDT data have already begun, spurring new
annotations that were not part of the original corpus design.  In
order to establish a better estimate of ASR performance on this data
set, NIST selected a random sample of 530 individual news stories from
the TDT2 English corpus, totaling 10 hours of speech.  The LDC adapted
the text for these stories to the Hub-4 transcription specifications
for broadcast news, and carefully went over each story, adding in the
missing words and disfluencies, correcting spelling errors (common in
closed-caption text), and adding time stamps to break long turns by
news announcers into manageable phrases.  This 10-hour set of careful
transcription is now available from the LDC.

Other directions for annotation of TDT have included the
identification of named entities, in support of the TREC project and
related research, and the identification of new information across a
sequence of stories on a given topic.

In a sense, each of these various annotations could be said to stand
on its own as a sample for modeling a particular characteristic of
language behavior or information flow.  But research tasks moving
along these various lines will tend to intersect, and it will be
important to know to what extent their respective annotations
intersect as well.  The same crew may be handling the annotations for,
e.g., first story detection, marking of new information, and named
entity extraction, but these tasks might not be carried out in unison
(indeed they typically will not), they might be applied to discrete
subsets of the corpus, and even if they do overlap on some sampling of
the data, it might not be immediately obvious how to integrate these
different annotations.  Needless to say, it would not seem improbable
that researchers could find the intersection of these annotations to
be of some value.

\begin{figure*}[t]
\fbox{\parbox{0.97\linewidth}{
\begin{center}
\epsfig{figure=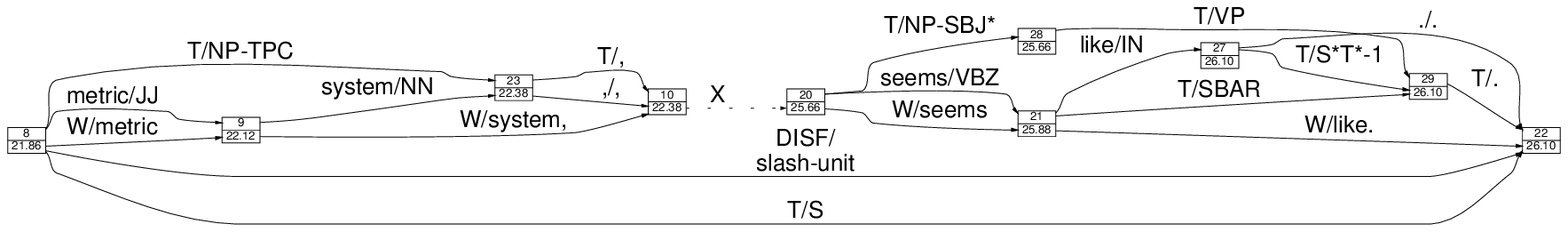,width=\linewidth}
\vspace*{2ex}
\epsfig{figure=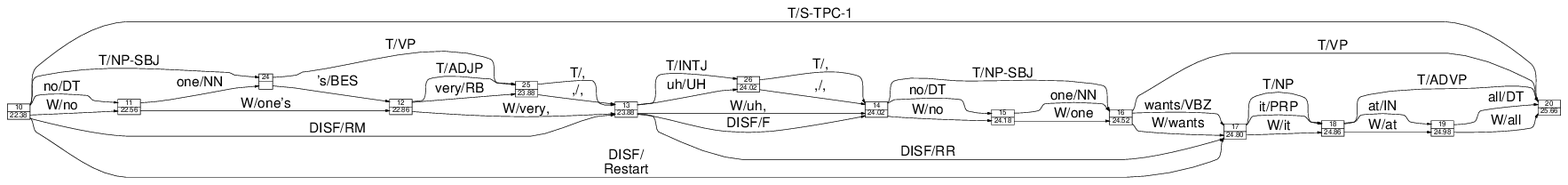,width=\linewidth}
\end{center}
}}
\vspace*{2ex}\hrule
\caption{Annotation Graph Fragment for Switchboard Data}
\label{fig:ag}
\end{figure*}

\section{Annotation Graphs as a Means of Integration}

Annotation graphs were introduced by Bird and Liberman
\shortcite{BirdLiberman99} as a convenient data model
which abstracts away from the many contingent details of
corpus file formats.  An important consequence of this
move, explored by Bird and Liberman \shortcite{BirdLiberman99dtag},
is that multiple independent annotations of a single corpus
can be accessed and analyzed simultaneously.  In this
section we discuss the case of SWB, and the
data shown in Figure~\ref{fig:swb}.

Figure~\ref{fig:ag} shows the annotation graph for this
SWB data, corresponding to
the interval [21.86, 26.10].  In this graph,
word arcs have type \smtt{W/},
Treebank arcs have \smtt{T/} and disfluency arcs
have \smtt{DISF/} type.  Types for the part-of-speech
arcs have been omitted for sake of clarity
(i.e. \smtt{Pos/metric/JJ} is written as just \smtt{metric/JJ}).
For readability, the graph is represented in two
pieces; the lower piece should be interpolated into the
upper piece at the position of the dotted arc labeled \smtt{X}.
Observe that the equivocation about
the tokenization of punctation from Figure~\ref{fig:swb}
is preserved in the annotation graph.

The ability to merge diverse layers of independent annotations into a
single graph derives from the definition of nodes for connecting the
arcs.  The nodes are anchored at specific points along a time line
representing the signal being annotated.  In SWB, the LDC
interim transcripts, the ICSI phonetic transcripts, and the ISIP
revised transcripts each represent a distinct segmentation of the time
lines for the corresponding signals.  The migration of these data sets
into annotation graphs provides a stable, time-anchored basis for
cross-reference; owing to the divergent turn- or phrase-level
segmentations provided by each data set, the time line is in fact the
only practical basis for cross-reference.

Even though the phonetic transcripts cover only sparse portions of
most calls, these partial annotations for any one file can be mapped
coherently onto a single complete graph that includes the other
transcripts in their entirety, providing a well-defined algorithmic
approach for locating and resolving discrepancies in lexical content,
phrasal segmentation, and word-level time alignments.  The disfluency,
discourse and Treebank annotations are derived from the LDC
transcripts, and as demonstrated above, they can also be incorporated
into the one graph, making it quite simple to identify the particular
elements in those annotations that will be affected by corrections to
the underlying transcripts.  It follows that the propagation of
transcription repairs through all levels of annotation becomes a
well-behaved and accountable process.

Apart from the obvious benefits to corpus maintenance, this approach
to handling annotations provides an important capability for
integrating the results of diverse annotation efforts in new research.
If and when a prosodic annotation of SWB becomes available, we can
readily envision the ability to study interactions among intonational
focus, discourse function, syntactic structure, and phonological
processes, simply by adding the one new layer to the existing network
of other annotations.

The application of annotation graphs to TDT is equally fruitful.  The
stability of reference to the basic units of TDT corpora (sample
files, news stories, word tokens) is already well established, owing
to the fact that the corpus creation effort has been tightly
centralized.  But due to the overall bulk of the data, most new
annotations, especially those requiring human judgment, are likely to
be limited to cover only portions of the collection.  Again, these
partial annotations can form coherent annotation graphs on their own,
and can be treated atomically or integrated with other graph
structures as needed.  

The issue of data formatting for creation, storage, distribution and
research use of annotations is an independent concern, orthogonal to
the use of annotation graphs as a framework for handling corpora.  The
arc-and-node structure can be rendered into (and retrieved from) a
very simple XML data stream, and it is equally possible to create
filters that can populate an annotation graph by reading any chosen
data format, without loss of information.  Filters could also be made
to create a chosen data format from an annotation graph, though it's
possible that some information in the graph would not be preserved in
the process.

\section{Conclusion}

We have presented an overview of two large speech corpora, both of
which have received a wide range of divergent and independent
annotations.  For Switchboard, we have discussed some details about
the comparability and compatibility of the various annotations, and
have presented an analysis framework that will enable a high degree of
integration among them, in terms of both maintenance and research
use.  The case of TDT demonstrates that even with a centralized corpus
creation effort, there can still be problems with handling data
repairs and consistency when distinct sets of annotation must be
carried out simultaneously.  Both corpora present the need to
accommodate sparse annotations in a manner that does not sacrifice the
overall coherence of the larger corpus.  Annotation graphs provide an
effective framework for meeting this need.


\bibliographystyle{lrec2000}
\bibliography{multiuse} 

\end{document}